\documentclass[12pt,doublecolumn]{IEEEtran}
\usepackage{amsmath,amsthm,amsfonts,amssymb,bm,mathrsfs}
\usepackage{graphicx,subfigure}
\usepackage{multicol}
\usepackage[square, comma, sort&compress, numbers]{natbib}
\usepackage[colorlinks, linkcolor=black, citecolor=black, urlcolor=black]{hyperref}
\usepackage{mathrsfs}
\usepackage{algorithm}
\usepackage{algorithmic}
\usepackage{multirow,booktabs,threeparttable}
\usepackage{tablefootnote}
\usepackage{color}
\newcommand{\subparagraph}{}
\usepackage{verbatim}
\usepackage{caption}
\usepackage{url}
\usepackage[perpage,symbol*]{footmisc}
\usepackage{titlesec}
\usepackage{array}

\theoremstyle{remark}

\theoremstyle{plain}

\begin{document}

\title{Evaluation Mechanism of Collective Intelligence for Heterogeneous Agents Group}

\author{
Anna Dai, Zhifeng Zhao, Honggang Zhang, Rongpeng Li, Yugeng Zhou

\thanks{A. Dai, Z. Zhao, H. Zhang and R. Li are with Zhejiang University, Hangzhou 310027, China, (email: \{annadai, zhaozf, honggangzhang, lirongpeng\}@zju.edu.cn).}

\thanks{Y. Zhou is with Zhejiang Wanfeng Technology Development Company Limited, China, (email:yugeng.zhou@wfjyjt.com).}


}

\maketitle

\begin{abstract}
Collective intelligence is manifested when multiple agents coherently work in observation, interaction, decision-making and action. In this paper, we define and quantify the intelligence level of heterogeneous agents group with the improved \emph{Anytime Universal Intelligence Test(AUIT)}, based on an extension of the existing evaluation of homogeneous agents group. The relationship of intelligence level with agents composition, group size, spatial complexity and testing time is analyzed. The intelligence level of heterogeneous agents groups is compared with the homogeneous ones to analyze the effects of heterogeneity on collective intelligence. Our work will help to understand the essence of collective intelligence more deeply and reveal the effect of various key factors on group intelligence level.
\end{abstract}


\IEEEpeerreviewmaketitle

\section{Introduction}
Collective or group is a very common organizational structure of intelligent creatures. Collective intelligence means that group of individuals acting collectively in ways that seem intelligent\citep{handbook}.
And it only occurs when there are interactions between agents in the group\citep{Weschsler1971Concept}. 
A human group’s performance on a wide variety of tasks can be explained by a general collective intelligence factor.\citep{Anita2010Evidence}
The theory of collective intelligence is helpful for understanding many aspects of group performance, bringing benefits to scientific research and practical applications\citep{Woolley2015Collective}.

There are several authoritative methods to quantify the intelligence of isolated agent, but it's hard to quantify the intelligence of groups.
David L Dowe\citep{Dowe1998NonBehavioural} proposed an additional computational requirement on intelligence, the ability of expression, as an extention to the Turing Test.
C-Test is a test for comprehending ability, equally applicable to both humans and machines, which was presented by J Hernandez-Orallo\citep{Hernandez2000Beyond}.
 Kannan and Parker\citep{Kannan2007Metrics} proposed an effective metric for the evaluation of learning capability. They attempt to evaluate the quality of learning towards understanding system level fault-tolerance. Schreiner\citep{Schreiner2000Measuring} presented a study related to creating standard measures for systems that can be considered intelligent, which is realized by the US National Institute of Standards and Technology (NIST). 
 Javier Insa-Cabrera\citep{insa2012influence}\citep{insa2015instrumental} analysed the influence of including agents of different degrees of intelligence and identified the components that should be considered when measuring social intelligence in multi-agent systems. 
Presented by Fox and Martin\citep{Fox2003Understanding}, an agent benchmark model is developed as a basis for analyzing and comparing multiple agent systems with cognitive capabilities.
In the research of Anthon and Jannett\citep{Anthony2007Measuring}, the agent-based systems’ intelligence is based on the task intelligent costs. 
Hibbard\citep{Hibbard2011Measuring} proposed a metric for intelligence measuring based on a hierarchy of increasingly complex environment sets. And an agent’s intelligence is measured as the ordinal of the most difficult set of environments it can pass. 
Chmait et al.\citep{chmait2017understanding} proposed a metric considered “universal” and appropriate to empirically measure the intelligence level of different agents or groups. 
J Hernández-Orallo\citep{Hern2015On} presented a way to estimate the difficulty and discriminating power of any task instance. 
A measure for machine intelligence was proposed by Legg and Hutter\citep{Shane2006A}, who mathematically formalized essential features about human intelligence to produce a general measure of intelligence for arbitrary machines.

Nader Chmait\citep{chmait2017understanding,Chmait2015Observation,chmait2015measuring} provided an information-theoretic solution and at first time quantified and analyzed the impact of communication and observation abilities on the intelligence of homogeneous multi-agent system. They considered a series of factors hindering and influencing the effectiveness of interactive cognitive systems\citep{chmait2016factors,chmait2016dynamic,chmait2017information}.

Heterogeneous groups are the aggregations of two or more interactive agents of different behaviors\citep{chmait2016dynamic}. 
Heterogeneous agents have been used to study real-world problems regarding cybersecurity\citep{Grossklags2011Security} and economy\citep{Burlando2005Heterogeneous}.  
The intelligence level of heterogeneous groups can’t be achieved directly by homogeneous group model\citep{magg2007pattern}. In this paper we develop the mechanism of quantifying the intelligence level of heterogeneous group. We find that the intelligence level of heterogeneous collectives is higher than same size homogeneous collectives in most cases. And the composition/heterogeneity of heterogeneous collectives also has an important impact on the intelligence level.

The remainder of the paper is organized as follows. Section \ref{sec:Model&Method} introduces the model and mechanism we define.
The experiment settings and parameters are in Section \ref{sec:parameters}. We present our experiment results in Section \ref{sec:experiments&results} along with some discussion and analysis of quantitative results. In Section \ref{sec:conclusion}, we briefly draw conclusion and introduce the future directions.

\section{Model and Method}
\label{sec:Model&Method}
\emph{The Anytime Universal Intelligence Test (AUIT)}\citep{hernandez2010measuring} is a method to evaluate the intelligence level of homogeneous multi-agent groups. The test simulates agents working in a finite environment and calculate the rewards corresponding to their actions. The average rewards over all agents are considered as the intelligence level of this group. The model works in a toroidal grid space, named $The\ \Lambda ^{*}(Lambda\ Star) Environment$. 

\begin{figure}[!t]
	\centering
	\includegraphics[width=0.45\textwidth]{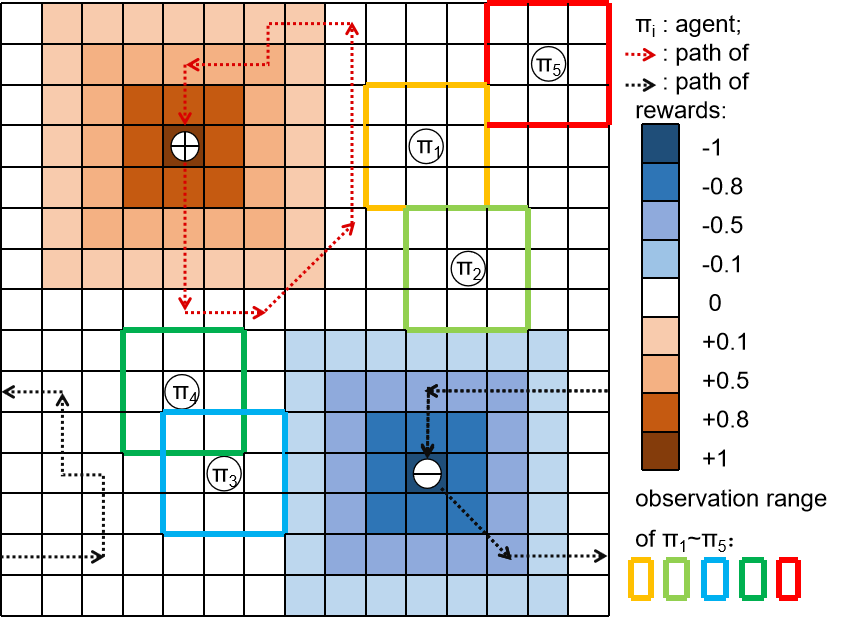}
	\captionsetup{font={scriptsize}}
	\caption{Test in a 15$\times$15 environment with 5 agents and 2 special objects.}
	\label{fig:testmodel}
\end{figure}

To evaluate the intelligence level of heterogeneous groups, we extend the model as shown in Fig. \ref{fig:testmodel}.
The environment is a toroidal grid space (periodic boundaries) which means that moving off one border makes you appear on the facing one. In this test environment, there are objects from finite set $\Omega=\left\{  \pi_1, \pi_2, ... , \pi_x, \oplus, \ominus   \right\} $ which contains working agents ($\Pi\subseteq\Omega, \Pi=\left\{ \pi_1, \pi_2, ... , \pi_x\right\} $) and two moving special objects, Good ($\oplus$) and Evil ($\ominus$). The two special objects travel in the environment with measurable complexity movement patterns. Each element in $\Omega$ can work as a finite set of move actions $A=\left\{\right.left, right, up, down, up-left, up-right, down-left, down-right, stay\left . \right\}$. Reward is defined as a function of the distance of the evaluated agent to objects $\oplus$ and $\ominus$\citep{chmait2016dynamic}. In the test environment, given an agent $\pi_j$, its reward $r_j$ is a real number, calculated as follows, in which $r_{j+}$ represents the reward caused by $\oplus$, and $r_{j-}$ represents the reward caused by $\ominus$:

\begin{equation}
\label{eq:r+}
r_{j+}=\left\{
\begin{array}{ll}
+1 & {d_{\pi_j,\oplus} =0}， \\
+0.8 & {d_{\pi_j,\oplus} =1}， \\
+0.5 & {d_{\pi_j,\oplus} =2}， \\
+0.1 & {d_{\pi_j,\oplus} =3}， \\
\quad0 & {\text{otherwise}}.
\end{array} \right.
\end{equation}

\begin{equation}
\label{eq:r-}
r_{j-}=\left\{
\begin{array}{ll}
-1 & {d_{\pi_j,\ominus} =0}， \\
-0.8 & {d_{\pi_j,\ominus} =1}， \\
-0.5 & {d_{\pi_j,\ominus} =2}， \\
-0.1 & {d_{\pi_j,\ominus} =3}， \\
\quad0 & {\text{otherwise}}.
\end{array} \right.
\end{equation}

\begin{equation}
\label{eq:r_sum}
r_j=r_{j+}+r_{j-}
\end{equation} 

\begin{figure}[!t]
	\centering
	\includegraphics[width=0.45\textwidth]{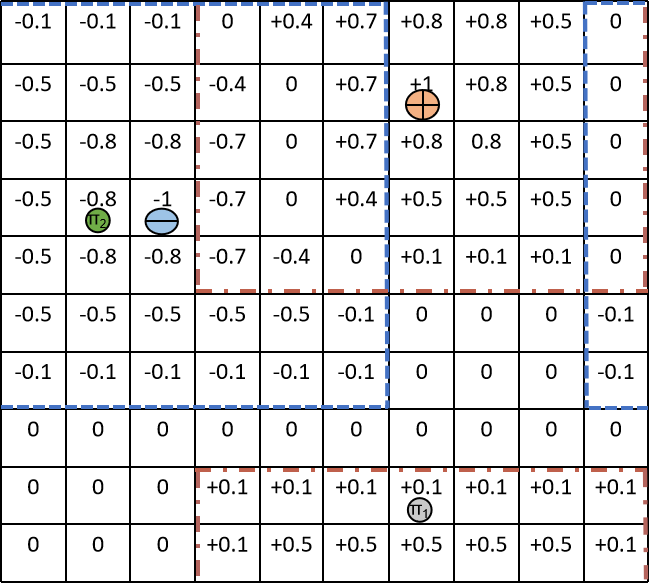}
	\captionsetup{font={scriptsize}}
	\caption{The snapshot of reward map.}
	\label{fig:environment}
\end{figure}

$d_{\pi_j,\oplus}$ and $d_{\pi_j,\ominus}$ each means the (toroidal) chessboard distance\citep{lee1996chessboard} between $\pi_j$ and $\oplus$ or $\ominus$, e.g. in a 10-by-10 grid-world, the distance from cell (2, 1) to (2, 10) is 1. The reward of agent is the combination of the effects of two distances. The  snapshot of agent rewards map is shown in Fig. \ref{fig:environment}.

The two special objects act as the moving targets in the evaluation, and the agents' work is to chase Good ($\oplus$) and keep away from Evil ($\ominus$). With these settings, a test episode consist of a series of $\vartheta$ iterations works as Algorithm~\ref{alg1}.

\begin{algorithm}\small
	\caption{Evaluation algorithm} \label{alg1}
	\begin{algorithmic}[1]
		\REQUIRE $\Pi$(set of n evaluated heterogeneous agents), special objects($\oplus$ and $\ominus$), $A$(set of actions), environment size m$\times$m, iteration number $\vartheta$ 
		\ENSURE The evaluation of an n-agent group’s intelligence 
		\STATE Agents from $\Pi\subseteq\Omega$ and the two special objects $\oplus$ and $\ominus$ are randomly distributed in the m-by-m toroidal grid-world $//$ Initialize
		\FOR{$i \gets 1$ to $\vartheta$} 
		\FOR{$j \gets 1$ to $n$}
		\STATE The environment sends an observation to $\pi_j^i$ \\
		$//$ Observation
		\ENDFOR
		\FOR{$j \gets 1$ to $n$}
		\STATE $\pi_j^i$ interacts with other agents about the observation and takes an action from $A$ 
		$//$ Action
		\ENDFOR
		\STATE The two special objects $\oplus$ and $\ominus$ perform the next action in their movement pattern and renew the rewards distribution in the environment
		\FOR{$j \gets 1$ to $n$}
		\STATE The environment returns a reward $r_j^i$ to $\pi_j^i$ according to its distance to the special objects $//$ Reward
		\ENDFOR
		\ENDFOR 
		\STATE Return $\tilde{R}_{\Pi,\mu,\vartheta} \gets  \frac{\sum_{j=1}^{n}\sum_{i=1}^{\vartheta}r_j^i}{n\times\vartheta}$	
	\end{algorithmic}
\end{algorithm}	

In Algorithm \ref{alg1}, an observation means the reward information of $\pi_j^i$’s observation range (1 Moore neighbour cells)\citep{gray2003mathematician}. The evaluation result is the average reward of each agent over each iteration\citep{chmait2016dynamic}, shown in Algorithm \ref{alg1}. 

In environment $\mu$, there are two types of complexity. One is task complexity $K(\mu)$\citep{chmait2016dynamic}, which corresponds to the difficulty of the task. And it is represented by the Kolmogorov complexity\citep{vitanyi1997introduction} of the two special objects’ movement patterns. The other one is search space complexity or environmental complexity $H(\mu)$\citep{chmait2016dynamic}, represented by Shannon entropy\citep{shannon1948mathematical} of the environment, which stands for the uncertainty of $\mu$ and corresponds to the size of the environment. To evaluate the collectives over the same task complexity, we do the simulation with the special objects following the same movement pattern while the other settings (total number of agents, environment size, test and iteration times) remain the same. And we increase the search space complexity by enlarging the size of the environment.

In the heterogeneous group evaluation, we take several types of agent into consideration. They include Local Search Agent, Oracle Agent, 
and Random Agent.

\begin{itemize}
	\item [a.] Local Search Agent: This kind of agent will choose the cell of highest reward in its observation range to be the target cell in one iteration. If the rewards of cells in the observation range are all equal, it will randomly choose one.
	\item [b.] Oracle Agent: It knows the movement pattern of the Good special object and can get close to it in the fastest way. 
	\item [c.] Random Agent: It randomly chooses one neighbor cell as its next moving target cell.
\end{itemize}

The synergy among the agents in the group is crucial to the performance\citep{liemhetcharat2014weighted}. The evaluation also considers different group communication methods, such as Talking, Stigmergy, Imitation. 

\begin{itemize}
	\item [a.] Talking(direct communication) : Agents exchange observation information with each other in communication range, then choose the highest reward cell as their moving target.
	\item [b.] Stigmergy\citep{grasse1959reconstruction}(indirect communication) : Agents get their own observation exactly. Each of them gets others' observations with random fake rewards. 
	\item [c.] Imitation: Agents take the same action as other agent in their observation range. When there are more than one agent in range, they randomly choose one to follow. 
\end{itemize}

In each iteration, different agents get information according to the communication methods they take. Agents with direct communication will get all agents' exact observations, while agents with indirect communication will get others' observations with bias. And agents with imitation will get the action information of the other agents in their observation range. After that, agents choose and perform actions. Then they get rewards from the environment. At the end of a test episode, the average rewards for each agent are considered as the intelligence level of the group. The test environment is like a real map, while the agent is like UAV(unmanned aerial vehicle). Based on the observation of the environment, different communication methods and decentralized groups, UAVs gather for search and avoidance missions, providing the test method with a practical application scenario.

\section{Numerical Experiment Parameters}
\label{sec:parameters}
Our experiments try to figure out the impact of agent composition, group size, environmental complexity and evaluation time on the intelligence level of heterogeneous groups. And we also compare the intelligence of heterogeneous group with same size homogeneous group to understand the impact of heterogeneity. The agents and communication methods we use in our simulation, as well as the corresponding symbols are listed in Table~\ref{tab:symbols}.

\begin{table}[h] 
	\caption{Symbols and Expression} 
	\centering 
	\scriptsize
	\label{tab:symbols}
	\renewcommand\arraystretch{1.5}
	\begin{tabular}{m{1.2cm}<{\centering} m{5.8cm}<{\centering}} 
		\hline  
		\hline
		{\bf \small Symbol} & {\qquad {\bf\small Meaning}}  \\ 
		\hline 
		{$SL$} & Local search agents using indirect communication (Stigmergy Local Search Agent). \\  
		\hline
		{$TL$} & Local search agents using direct communication (Talking Local Search Agent) \\ 
		\hline  
		{$IL$} & Local search agents follow any other type of agent in its observation range (Imitative Local Search Agent), or just work as a Local Search agent exchanging no information if no agent in observation range.\\
		\hline  
		$O$ & Oracle agents. They offer their observations according to other agents' communication method.\\
		\hline  
		$R$ & Random agents. They offer their observations according to other agents' communication method.\\
		\hline  
		{$\&$} & “$Ax\&By$” represents a heterogeneous group consists of $x$ type $A$ agents and $y$ type $B$ agents ($x,y\in N$)\\      
		\hline  
		\hline  
	\end{tabular}\\[1ex]  
	e.g..“$SL10$” represents a homogeneous group consists of 10 $SL$s.
	
	“$SL9\&O1$” represents a heterogeneous group consists of 9 $SL$s and 1 $O$.
\end{table}

In the evaluation of heterogeneous groups, we mainly carry out the simulation in a 20$\times$20 environment, corresponding to $H(\mu)$=17.2bits. And each test contains 20 iterations. When we need to figure out the impact of environmental complexity, time and group size, $H(\mu)$ varies from 13.2bits to 19.6bits, number of iterations varies from 10 to 500, and agent number varies from 10 to 60(the ratio of the components remains the same). 
In the comparison of heterogeneous groups and homogeneous groups, we use same size homogeneous groups with other settings all the same.

\section{Numerical Experiments and Results}   
\label{sec:experiments&results}

\subsection{Evaluation of heterogeneous agents group}
\label{sec:hetegroup}
\subsubsection{Agents Composition}
\label{sec:hete_composition}
\begin{figure}[tbh]
	\centering
	\includegraphics[scale=0.45]{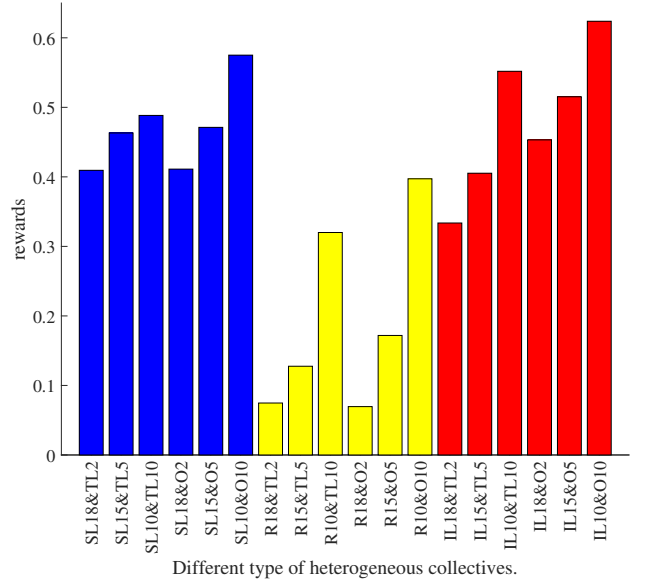}
	\caption{Evaluation of heterogeneous collectives.} \label{fig:hete_composition}
\end{figure}

In heterogeneous groups, agents of different decision strategies or communication methods work cooperatively. We combine different agents to get heterogeneous groups and evaluate their intelligence, showing the results in Fig.~\ref{fig:hete_composition}. The intelligence level of heterogeneous groups is mainly determined by the intelligence level of the components. And the same size group shows higher intelligence level as the heterogeneity gets stronger, which indicates that heterogeneity can help improve the group intelligence.

\subsubsection{Impact of Agent Number}
\label{sec:hete_number}
\begin{figure}[tbh]
	\centering
	\includegraphics[scale=0.5]{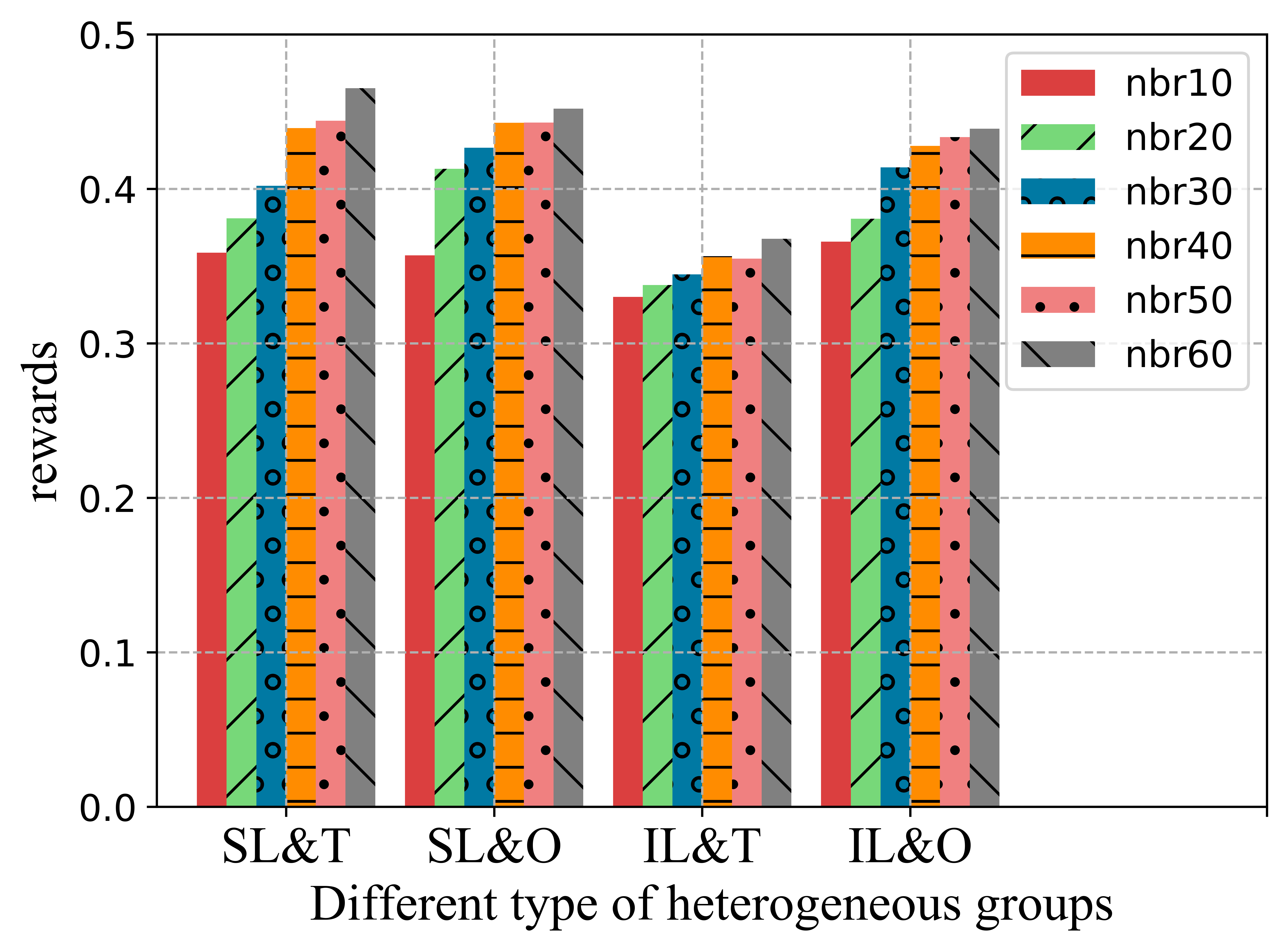}
	\caption{Evaluation of heterogeneous collectives of different agent numbers.} \label{fig:hete_number}
\end{figure}

Just as the saying goes ``many hands make light job''. We enlarge the group size step by step to observe the changes in their quantified collective intelligence level. 
In Fig.~\ref{fig:hete_number}, the ratio of one type agents to the other type in the collectives remains 9:1(e.g., the second bar of the first cluster with ``$/$'' pattern represents a collective consists of 18 $SL$s and 2 $TL$s). 
In the test environment, more agents means more information can be observed, leading to a better reward. With number of agents continue increasing, the result will come to an upper limit.

\subsubsection{Impact of Environment Complexity}
\label{sec:hete_Hu}
\begin{figure}[tbh]
	\centering
	\includegraphics[scale=0.5]{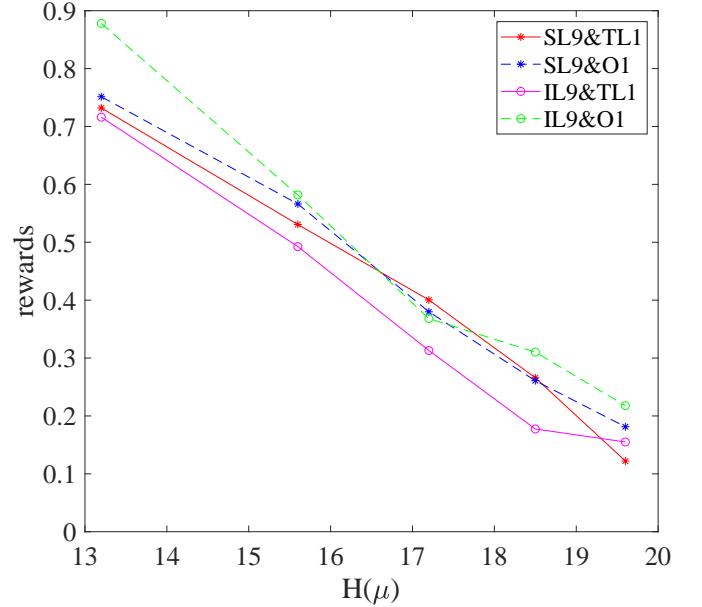}
	\caption{Evaluation of heterogeneous collectives with different environment complexities.} \label{fig:hete_Hu}
\end{figure}

We gradually increase the size of the environment to increase the search space complexity. Fig. \ref{fig:hete_Hu} illustrates that heterogeneous collective intelligence will decrease with the environmental complexity increasing. With the environment space getting larger, it is much harder to meet $\oplus$ and elude $\ominus$ in finite time, resulting in the decrease of heterogeneous group performance. When the environment is too large for any agent to sense or learn the position of special objects, the group may just perform like all agents walking randomly and aimlessly, and they get rewards close to 0. 

\subsubsection{Impact of Evaluation Time}
\label{sec:hete_time}
\begin{figure}[tbh]
	\centering
	\includegraphics[scale=0.47]{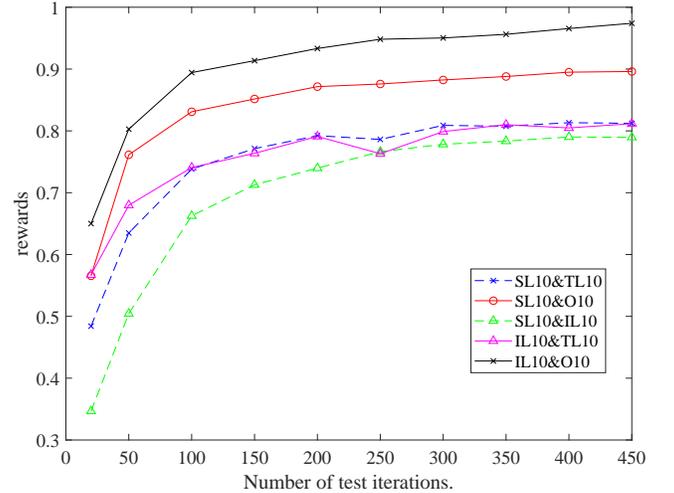}
	\caption{Evaluation of heterogeneous collectives with different evaluation time.} \label{fig:hete_time}
\end{figure}

We extend the evaluation time by increasing the number of iterations for each test. Then we get Fig.~\ref{fig:hete_time}, in which heterogeneous collectives show performance increase. As time goes longer, agents get more chance to seek and follow $\oplus$ as well as staying away from $\ominus$. And the evaluation results gradually come to an upper limit with test time long enough. 
In Fig.~\ref{fig:hete_time}, the gap between $IL10\&TL10$ and $IL10\&O10$ is larger than that between $SL10\&TL10$ and $SL10\&O10$. That means the intelligence level of indirect communicating heterogeneous groups is more stable than that of imitative heterogeneous groups. 

\subsection{Comparison between Homogeneous Agents Group and Heterogeneous Ones}
\label{sec:compare}
\subsubsection{Agents Composition}
\label{sec:comp_composition}
\begin{figure}[tbh]
	\centering
	\includegraphics[scale=0.37]{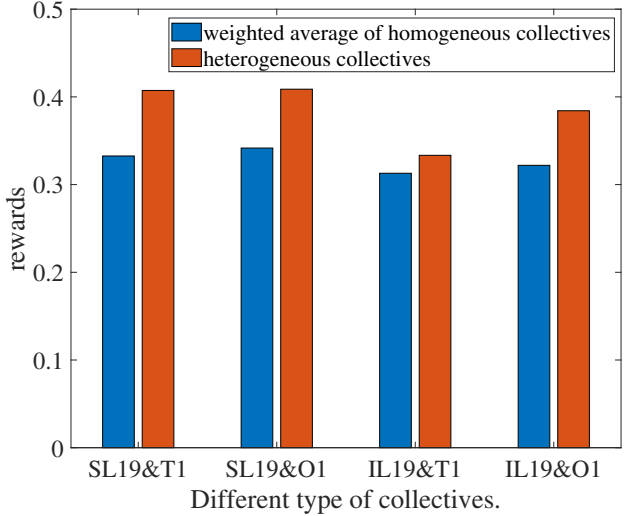}
	\caption{Comparison of weighted average of homogeneous collectives and heterogeneous collectives.} \label{fig:comp_composition}
\end{figure}

We evaluate the intelligence level of heterogeneous groups and same size homogeneous groups, 
then we calculate the weighted average of homogeneous groups according to the composition of the heterogeneous groups, and we get Fig.~\ref{fig:comp_composition}. For example, ``SL19$\&$T1'' of ``weighted average of homogeneous collectives'' is calculated as
\begin{equation}
\label{eq:comp}
(19\times SL_{homo}+1\times T_{homo})/20
\end{equation}
in which $SL_{homo}$ is the intelligence level of 20-agent homogeneous SL group, and $T_{homo}$ is the intelligence level of 20-agent homogeneous T group. 
The heterogeneous group intelligence level is apparently higher than the avarage level of the components when they are in the homogeneous groups. And it implies that heterogeneity does have a positive impact on group intelligence level.  


\begin{figure}[tbh]
	\centering
	\includegraphics[scale=0.37]{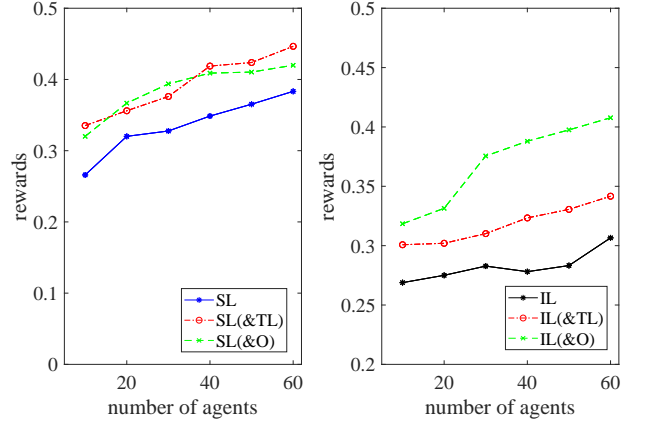}
	\caption{Comparison between homogeneous groups and the less intelligent part of agents in heterogeneous groups with agent number increasing.} \label{fig:nbrcontra}
\end{figure}

In Fig. \ref{fig:nbrcontra}, “$SL(\&TL)$” means the quantified result of $SL$ in the heterogeneous collectives which also contain $TL$. In these heterogeneous collectives, the ratio of $SL$ or $IL$ to the other type of agents remains 9:1. In Fig. \ref{fig:nbrcontra}, the homogeneous part in heterogeneous collectives apparently shows the improvement caused by heterogeneity. In the left part of Fig. \ref{fig:nbrcontra}, the improvement caused by $TL$ is similar to that caused by $O$. While in the right part, working together with $O$ can make better improvement for the performance of $IL$, since they are highly dependent on other type of agents.

\subsubsection{Impact of Agent Number}
\label{sec:comp_number}
\begin{figure}[tbh]
	\centering
	\includegraphics[scale=0.52]{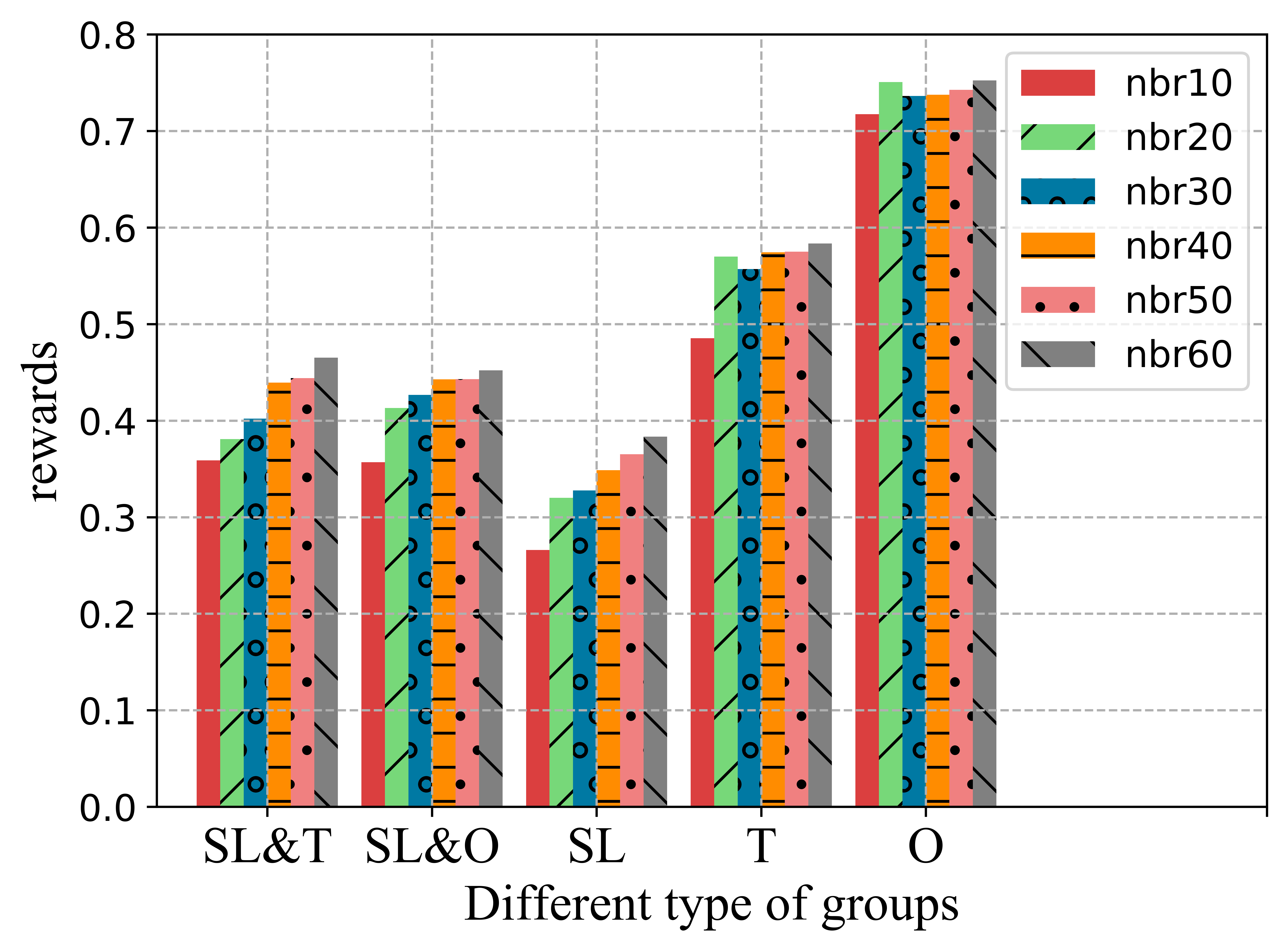}
	\caption{Evaluation of groups of various agent numbers.} \label{fig:comp_nbr}
\end{figure}

The group performance going up with the agents number increasing in both heterogeneous and homogeneous cases. In Fig. \ref{fig:comp_nbr}, homogeneous group “$TL$” is apparently less intelligent than “$O$”. However, in heterogeneous occasions, “$SL\&TL$” and “$SL\&O$” get similiar performances, indicating that in indirect communication the improvement caused by heterogeneity may have no absolute relation to the original performance of the wiser agents.
Indirect communicating groups improve to a similar intelligence level, even if agents of different intelligence level are added to the groups. 

\subsubsection{Impact of Environment Complexity}
\label{sec:comp_Hu}
\begin{figure}[tbh]
	\centering
	\includegraphics[scale=0.3]{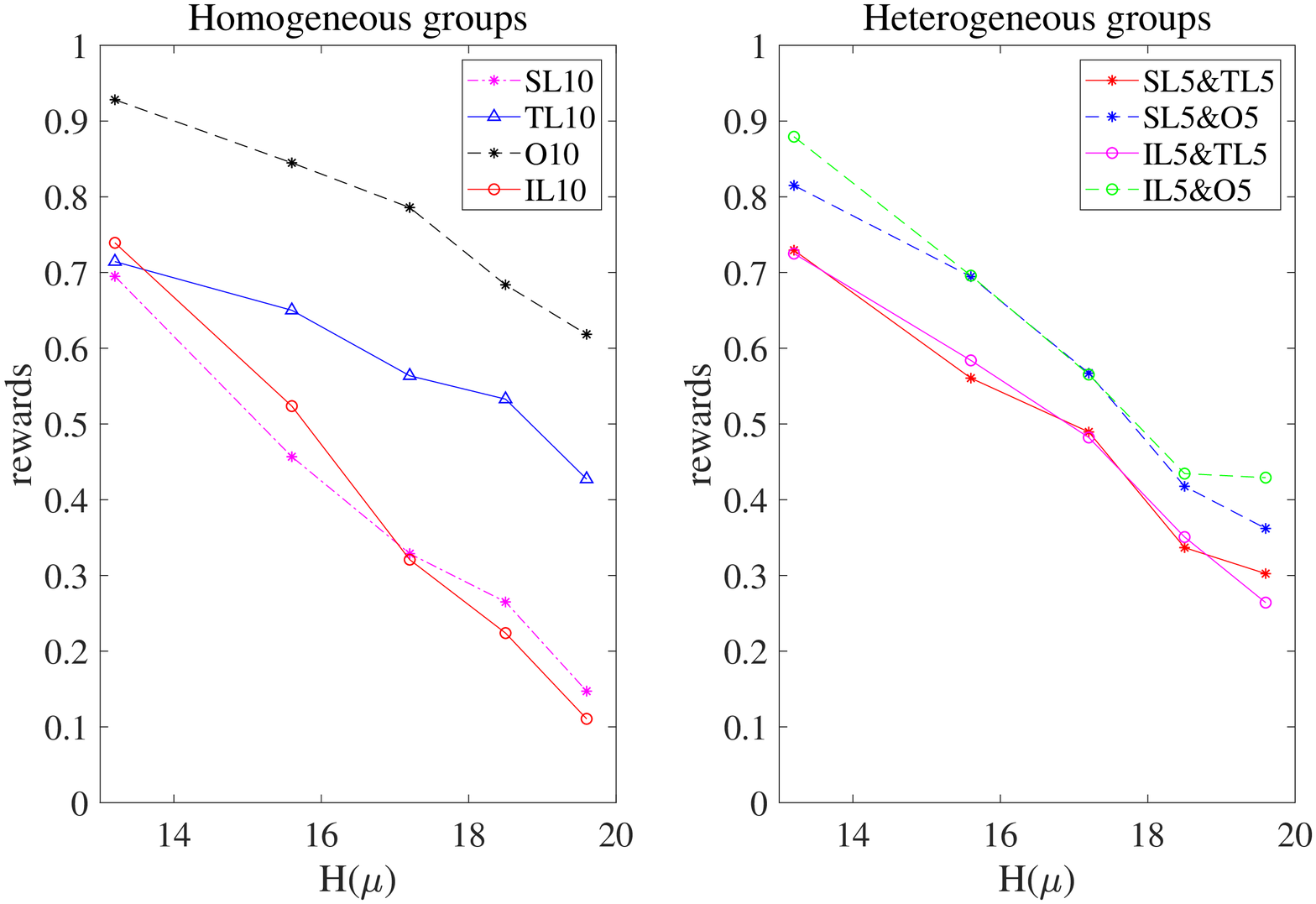}
	\caption{Evaluation of homogeneous and heterogeneous groups with different environment complexities.} \label{fig:comp_Hu}
\end{figure}

In Fig.~\ref{fig:comp_Hu}, homogeneous groups and heterogeneous groups both show intelligence level decrease when environmental complexities increase. The decrease speed seems to be the average of the components. The impact of environment complexity on heterogeneous group intelligence level is mostly determined by the components. The heterogeneity can make the group performance more stable. 

\subsubsection{Impact of Evaluation Time}
\label{sec:comp_time}
\begin{figure}[tbh]
	\centering
	\includegraphics[scale=0.25]{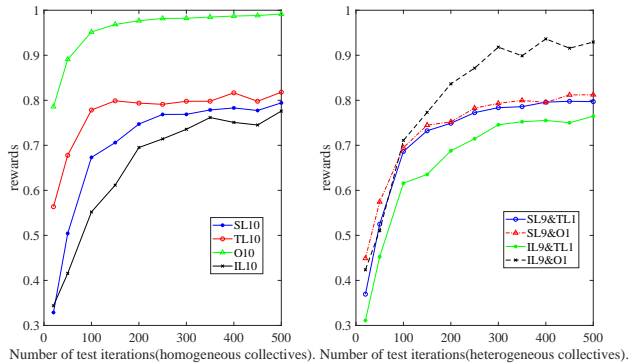}
	\caption{Evaluation of of different evaluation time.} \label{fig:comp_time}
\end{figure}

Homogeneous and heterogeneous collectives all show similar performance increase with time expanding, shown in Fig.~\ref{fig:comp_time}. 
Most heterogeneous groups' performance rise slower than that of homogeneous groups. It is quite obvious that “$IL9\&O1$” outperforms “$TL10$” when time is long enough. And “$SL9\&O1$” may work as well as “$TL10$”. The performance of heterogeneous collectives which contain few high-performance agents is getting very close to that of the homogeneous collectives contain only high-performance agents. 
When time expanding, we can expect heterogeneous groups mainly consist of low-performance agents to get high intelligence level. That is of great significance in reality since high-performance agents, like $TL$ and $O$, are often energy-intensive or even impractical.

\section{Conclusion and Future Works}
\label{sec:conclusion}
To evaluate the intelligence level of heterogeneous group, we improve The Anytime Universal Intelligence Test(AUIT) model and method, which can be applied to real UAV mission scenarios. We evaluate the intelligence level of different heterogeneous groups and study the impact of agent composion together with communication methods, group size, environment complexity and evaluation time. Experiment results prove that (a) Heterogeneity can improve the group intelligence level; (b)More agents and longer test time can also lead to better group performance; (c)The intelligence level improvement of heterogeneous groups that mainly adopt indirect communication is quite stable, while groups most made of imitative agents are more likely to be affected by external conditions such as the space size and evaluating time. 

In the future, to make the simulation closer to the actual situation, especially for the indirect communication method, the generation of fake rewards should have something to do with the distances between agents. To expand our work, we consider enriching the agent types, for example, incorporating reinforcement learning agents, and adopting different agent organizational structures in simulation. 

\renewcommand\refname{\textsc{Reference}}

\bibliographystyle{IEEEtran}
\bibliography{evaluation}

\end{document}